\documentclass[11pt,a4paper]{article}
\usepackage{coling2018}
\usepackage{times}
\usepackage{url}
\usepackage{latexsym}
\usepackage{siunitx}
\usepackage{graphicx}
\usepackage{tabu}
\usepackage{multirow}
\usepackage{siunitx}

\title{Open Information Extraction on Scientific Text: An Evaluation}

\author{Paul Groth\\
  Elsevier Labs \\
  1600 \\
  John F. Kennedy Blvd.\\ 
  Suite 1800\\
  Philadelphia, PA\\
  {\tt \small p.groth@elsevier.com} \\\And
  Mike Lauruhn \\
  Elsevier Labs \\
  1600 \\
  John F. Kennedy Blvd.\\ 
  Suite 1800\\
  Philadelphia, PA\\
    {\tt \small m.lauruhn@elsevier.com}
  \\\And
  Antony Scerri \\
  Elsevier Labs \\
	1600 \\
  John F. Kennedy Blvd.\\ 
  Suite 1800\\
  Philadelphia, PA\\
    {\tt \small a.scerri@elsevier.com}  \\\And
  Ron Daniel, Jr.\\
  Elsevier Labs \\
 1600 \\
  John F. Kennedy Blvd.\\ 
  Suite 1800\\
  Philadelphia, PA\\
    {\tt \small r.daniel@elsevier.com}
  }

\date{}

\begin{document}
\maketitle
\begin{abstract}
Open Information Extraction (OIE) is the task of the unsupervised creation of structured information from text. OIE is often used as a starting point for a number of downstream tasks including knowledge base construction, relation extraction, and question answering. While OIE methods are targeted at being domain independent, they have been evaluated primarily on newspaper, encyclopedic or general web text. In this article, we evaluate the performance of OIE on scientific texts originating from 10 different disciplines. To do so, we use two state-of-the-art OIE systems using a crowd-sourcing approach. We find that OIE systems perform significantly worse on scientific text than encyclopedic text. We also provide an error analysis and suggest areas of work to reduce errors. Our corpus of sentences and judgments are made available.
\end{abstract}

\section{Introduction}
%
% The following footnote without marker is needed for the camera-ready
% version of the paper.
% Comment out the instructions (first text) and uncomment the 8 lines
% under "final paper" for your variant of English.
% 
\blfootnote{
    %
    % for review submission
    %
    %\hspace{-0.65cm}  % space normally used by the marker
    % % final paper: en-uk version 
    %
    % \hspace{-0.65cm}  % space normally used by the marker
    % This work is licenced under a Creative Commons 
    % Attribution 4.0 International Licence.
    % Licence details:
    % \url{http://creativecommons.org/licenses/by/4.0/}
    % 
    % % final paper: en-us version 
    %
     \hspace{-0.65cm}  % space normally used by the marker
     This work is licensed under a Creative Commons 
     Attribution 4.0 International License.
     License details:
     \url{http://creativecommons.org/licenses/by/4.0/}
}

The scientific literature is growing at a rapid rate~\cite{bornmann2015growth}. To make sense of this flood of literature, for example, to extract cancer pathways~\cite{distant-supervision-cancer-pathway-extraction-text} or find geological features~\cite{leveling2015tagging}, increasingly requires the application of natural language processing. Given the diversity of information and its constant flux, the use of unsupervised or distantly supervised techniques are of interest~\cite{distant-supervision-relation-extraction-beyond-sentence-boundary}. In this paper, we investigate one such unsupervised method, namely, Open Information Extraction (OIE)~\cite{Banko2007}. OIE is the task of the unsupervised creation of structured information from text. OIE is often used as a starting point for a number of downstream tasks including knowledge base construction, relation extraction, and question answering~\cite{Mausam2016}. 

While OIE has been applied to the scientific literature before~\cite{DBLP:conf/akbc/GrothPMAD16}, we have not found a systematic evaluation of OIE as applied to scientific publications. The most recent evaluations of OIE extraction tools~\cite{Gashteovski2017,Schneider2017} have instead looked at the performance of these tools on traditional NLP information sources (i.e. encyclopedic and news-wire text). Indeed, as~\cite{Schneider2017} noted, there is little work on the evaluation of OIE systems. Thus, the goal of this paper is to evaluate the performance of the state of the art in OIE systems on scientific text. 

Specifically, we aim to test two hypotheses:
\begin{enumerate}
\item H1: There is no significant difference in the accuracy of OIE systems on scientific vs. general audience content.
\item H2: There is no significant difference in performance of current state-of-the-art OIE systems on scientific and medical content.\end{enumerate}

Additionally, we seek to gain insight into the value of unsupervised approaches to information extraction and also provide information useful to implementors of these systems. We note that our evaluation differs from existing OIE evaluations in that we use crowd-sourcing annotations instead of expert annotators. This allows for a larger number of annotators to be used. All of our data, annotations and analyses are made openly available.\footnote{\url{http://dx.doi.org/10.17632/6m5dyx4b58.2}}

The rest of the paper is organized as follows. We begin with a discussion of existing evaluation approaches and then describe the OIE systems that we evaluated. We then proceed to describe the datasets used in the evaluation and the annotation process that was employed. This is followed by the results of the evaluation including an error analysis. Finally, we conclude.

\section{Existing Evaluation Approaches}
OIE systems analyze sentences and emit relations between one predicate and two or more arguments (e.g. Washington :: was :: president). The arguments and predicates are not fixed to a given domain. (Note, that throughout this paper we use the word `triple'' to refer interchangeably to  binary relations.) Existing evaluation approaches for OIE systems have primarily taken a ground truth-based approach. Human annotators analyze sentences and determine correct relations to be extracted. Systems are then evaluated with respect to the overlap or similarity of their extractions to the ground truth annotations, allowing the standard metrics of precision and recall to be reported. 

This seems sensible but is actually problematic because of different but equivalent representations of the information in an article. For example, consider the sentence ``The patient was treated with Emtricitabine, Etravirine, and Darunavir''. One possible extraction is:
\begin{quote}
(The patient :: was treated with :: Emtricitabine, Etravirine, and Darunavir)
\end{quote}
Another possible extraction is:
\begin{quote}
(The patient :: was treated with :: Emtricitabine)\\
(The patient :: was treated with :: Etravirine)\\
(The patient :: was treated with :: Darunavir)
\end{quote}
Neither of these is wrong, but by choosing one approach or the other a pre-constructed gold set will falsely penalize a system that uses the other approach.

From such evaluations and their own cross dataset evaluation, \cite{Schneider2017} list the following common errors committed by OIE systems: 

\begin{itemize} 
\item wrong boundaries around the arguments or predicate of a relation;
\item generation of redundant relations from the same sentence;
\item wrong extractions (e.g. omitting large parts of a sentence as an argument);
\item missing extractions - extractions that should have been extracted from a sentence;
\item uninformative extractions that omit critical information from a sentence;
\end{itemize}

In our evaluation, we take a different approach. We do not define ground truth relation extractions from the sentences in advance. Instead, we manually judge the correctness of each extraction after the fact. We feel that this is the crux of the information extraction challenge. Is what is being extracted correct or not?  This approach enables us to consider many more relations through the use of a crowd-sourced annotation process.  Our evaluation approach is similar to the qualitative analysis performed in \cite{Schneider2017} and the evaluation performed in ~\cite{Gashteovski2017}. However, our evaluation is able to use more judges (5 instead of 2) because we apply crowd sourcing.\footnote{\cite{Schneider2017} also perform a ground truth based evaluation. This evaluation had many more sentences than ours, but used predefined gold extractions from multiple corpora annotated according to different criteria; illustrating the kinds of issues we mention with such an evaluation method.} For our labelling instructions, we adapted those used by~\cite{Gashteovski2017} to the crowd sourcing setting.\footnote{Labelling instructions are included in the associated dataset. - \url{http://dx.doi.org/10.17632/6m5dyx4b58.2#file-7edb4b86-c0e6-4169-aea0-4862d39461d3}} 

As previously noted existing evaluations have also only looked at encyclopedic or newspaper corpora. Several systems (e.g.~\cite{Banko2007,Corro2013}) have looked at text from the web as well, however, as far as we know, none have specifically looked at evaluation for scientific and medical text.

\section{Systems}
We evaluate two OIE systems (i.e. extractors). The first, OpenIE 4 \cite{Mausam2016}, descends from two popular OIE systems OLLIE~\cite{Fader2011} and Reverb~\cite{Fader2011}. We view this as a baseline system. The second was MinIE~\cite{Gashteovski2017}, which is reported as performing better than OLLIE, ClauseIE~\cite{Corro2013} and Stanford OIE~\cite{Corro2013}. MinIE focuses on the notion of minimization - producing compact extractions from sentences. In our experience using OIE on scientific text, we have found that these systems often produce overly specific extractions that do not provide the redundancy useful for downstream tasks. Hence, we thought this was a useful package to explore.   

We note that both OpenIE 4 and MiniIE support relation extractions that go beyond binary tuples, supporting the extraction of n-ary relations. We note that the most recent version of Open IE (version 5) is focused on n-ary relations. For ease of judgement, we focused on binary relations. Additionally, both systems support the detection of negative relations. 

In terms of settings, we used the off the shelf settings for OpenIE 4. For MinIE, we used their ``safe mode" option, which uses slightly more aggressive minimization than the standard setting. In the recent evaluation of MiniIE, this setting  performed roughly on par with the default options~\cite{Gashteovski2017}. Driver code showing how we ran each system is available.\footnote{See directory "Code for applying information extraction tools" in the associated data - \url{http://dx.doi.org/10.17632/6m5dyx4b58.2#folder-39ce9705-ccf3-4f95-890a-508ce155ece4}}

\section{Datasets}
We used two different data sources in our evaluation. The first dataset (WIKI) was the same set of 200 sentences from Wikipedia used in \cite{Gashteovski2017}. These sentences were randomly selected by the creators of the dataset. This choice allows for a rough comparison between our results and theirs. 

The second dataset (SCI) was a set of 220 sentences from the scientific literature. We sourced the sentences from the OA-STM corpus.\footnote{\url{http://elsevierlabs.github.io/OA-STM-Corpus/}} This corpus is derived from the 10 most published in disciplines.  It includes 11 articles each from the following domains: agriculture, astronomy, biology, chemistry, computer science, earth science, engineering, materials science, math, and medicine. The article text is made freely available and the corpus provides both an XML and a simple text version of each article.  

We randomly selected 2 sentences with more than two words from each paper using the simple text version of the paper. We maintained the id of the source article and the line number for each sentence.   

\section{Annotation Process}
We employed the following annotation process. Each OIE extractor was applied to both datasets with the settings described above. This resulted in the generation of triples for 199 of the 200 WIKI sentences and 206 of the 220 SCI sentences. That is there were some sentences in which no triples were extracted. We discuss later the sentences in which no triples were extracted. In total 2247 triples were extracted.

The sentences and their corresponding triples were then divided. Each task contained 10 sentences and all of their unique corresponding triples from a particular OIE systems. Half of the ten sentences were randomly selected from SCI and the other half were randomly selected from WIKI.  Crowd workers were asked to mark whether a triple was correct, namely, did the triple reflect the consequence of the sentence. Examples of correct and incorrect triples were provided. Complete labelling instructions and the presentation of the HITS can be found with the dataset. All triples were labelled by at least 5 workers. 

Note, to ensure the every HIT had 10 sentences, some sentences were duplicated. Furthermore, we did not mandate that all workers complete all HITS. 

We followed recommended practices for the use of crowd sourcing in linguistics~\cite{Erlewine2016}. We used Amazon Mechanical Turk as a means to present the sentences and their corresponding triples to a crowd for annotation. Within Mechanical Turk tasks are called Human Intelligence Tasks (HITs). To begin, we collected a small set of sentences and triples with known correct answers. We did this by creating a series of internal HITs and loaded them the Mechanical Turk development environment called the Mechanical Turk Sandbox. The HITs were visible to a trusted group of colleagues who were asked to complete the HITs. 

Having an internal team of workers attempt HITs provides us with two valuable aspects of the eventual production HITs. First, internal users are able to provide feedback related to usability and clarity of the task. They were asked to read the instructions and let us know if there was anything that was unclear. After taking the HITs, they are able to ask questions about anomalies or confusing situations they encounter and allow us to determine if specific types of HITs are either not appropriate for the task or might need further explanation in the instructions. In addition to the internal users direct feedback, we were also able to use the Mechanical Turk Requester functionality to monitor how long (in minutes and seconds) it took each worker to complete each HIT. This would come into factor how we decided on how much to pay each Worker per HIT after they were made available to the public.

The second significant outcome from the internal annotations was the generation of a set of `expected' correct triples. Having a this set of annotations is an integral part of two aspects of our crowdsourcing process. First, it allows us to create a qualification HIT. A qualification HIT is a HIT that is made available to the public with the understanding the Workers will be evaluated based on how closely they matched the annotations of the internal annotators. Based upon this, the Workers with the most matches would be invited to work on additional tasks. Second, we are able to add the internal set of triples randomly amongst the other relations we were seeking to have annotated. This allows us to monitor quality of the individual Workers over the course of the project. Note, none of this data was used in the actual evaluation. It was only for the purposes of qualifying Workers. 

We are sensitive to issues that other researchers have in regards to Mechanical Turk Workers earning fair payment in exchange for their contributions to the HITs~\cite{Fort2011} . We used the time estimates from our internal annotation to price the task in order to be above US minimum wage. All workers were qualified before being issued tasks. Overall, we employed 10 crowd workers. On average it took 30 minutes for a worker to complete a HIT.  In line with \cite{10.1371/journal.pone.0057410}, we monitored for potential non-performance or spam by looking for long response times and consecutive submitted results. We saw no indicators of low quality responses. 

\section{Judgement Data and Inter-Annotator Agreement}
In total, 11262 judgements were obtained after running the annotation process. Every triple had at least 5 judgements from different annotators. All judgement data is made available.\footnote{`aggregated\_results\_anon.csv' in the associated dataset. - \url{http://dx.doi.org/10.17632/6m5dyx4b58.2#file-03de3c93-8a01-4cd2-bfe5-c5bfeaa4f492}} The proportion of overall agreement between annotators is 0.76 with a standard deviation of 0.25 on whether a triple is consequence of the given sentence. We also calculated inter-annotator agreement statistics. Using Krippendorff's  alpha inter-annotator agreement was 0.44. This calculation was performed over all data and annotators as Krippendorff's alpha is designed to account for missing data and work across more than two annotators. Additionally, Fleiss' Kappa and Scott's pi were calculated pairwise between all annotators where there were overlapping ratings (i.e. raters had rated at least one triple in common). The average Fleiss's Kappa was 0.41 and the average of Scott's pi was 0.37. Using~\cite{artstein2008inter} as a guide, we interpret these statistics as suggesting there is moderate agreement between annotators and that agreement is above random chance. This moderate level of agreement is to be expected as the task itself can be difficult and requires judgement from the annotators at the margin. 

Table \ref{hardtriples} shows examples of triples that were associated with higher disagreement between annotators. 
One can see for example, in the third example, that annotators might be confused by the use of a pronoun (him). Another example is in the last sentence of the table, where one can see that there might be disagreement on whether the subsequent prepositional phrase behind light microscope analysis should be included as part of the extracted triple. 

\begin{table}[h]
\begin{tabular}{l l p{4cm} l p{5cm}}
\textsc{System} & \textsc{Source} &  \textsc{Triple} & \textsc{Pairwise} & \textsc{Sentence} \\ 
& & & \textsc{Agreement} & \\
\hline
 MinIE & SCI & additional QUANT\_S\_1 on scalp face :: were digitized :: QUANT\_O\_1 anatomical landmarks & 40\% False & To coregister MEG and sMRI data, three anatomical landmarks (nasion and right and left preauriculars) as well as an additional 150+ points on the scalp and face were digitized for each subject using the Probe Position Identification (PPI) System (Polhemus, Colchester, VT). \\
 \\
OpenIE 4 & WIKI &  Dawlish :: to 2nd :: place L:in the Western League & 40\% True & \multirow{2}{5cm}{The previous season had seen him lead Dawlish to 2nd place in the Western League , their highest ever league finishing position .} \\
MiniIE & WIKI & him :: lead Dawlish to 2nd place in :: Western League' & 40\% False  \\
MinIE & SCI & $>\SI{250}{\micro\metre}$ fractions :: be found through :: light microscope analysis to be dominated by amorphous organic matter & 40\% True & 'The $>\SI{250}{\micro\metre}$ fractions, found through light microscope analysis to be dominated ($>90\%$) by amorphous organic matter (AOM), were also analysed for $\delta13C$.
 \end{tabular}
 \caption{Examples of difficult to judge triples and their associated sentences. \label{hardtriples}}
\end{table}

We take the variability of judgements  into account when using this data to compute the performance of the two extraction tools.  Hence, to make assessments as to whether a triple correctly reflects the content from which it is extracted, we rely on the unanimous positive agreement between crowd workers. That is to say that if we have 100\% inter-annotator agreement that a triple was correctly extracted we label it as correct.
 
\section{Experimental Results}

Table \ref{aggresults} show the results for the combinations of systems and data sources. The \textsc{Correct Triples} column contains the number of triples that are labelled as being correct by all annotators. \textsc{Total Triples} are the total number of triples extracted by the given systems over the specified data. Precision is calculated as typical where Correct Triples are treated as true positives. On average, 3.1 triples were extracted per sentence. 

\begin{table}
\begin{center}
\begin{tabular}{l l r r S[table-format=3.2]}
\textsc{System} & \textsc{Source} & \textsc{Total Triples} & \textsc{Correct Triples} & \textsc{{Precision}}  \\
\hline
OpenIE 4 + MinIE & SCI + WIKI & 2247 & 985 & 0.44 \\
OpenIE 4 + MinIE & WIKI & 1101 & 590 & 0.54\\
OpenIE 4 + MinIE & SCI &  1146 & 395 & 0.34\\
MinIE  & SCI + WIKI & 1591 & 617 & 0.39\\
MinIE & WIKI &  785 & 382 & 0.49\\
MinIE & SCI & 806 & 235 & 0.29 \\
OpenIE 4 & SCI + WIKI &  656 & 368 & 0.56 \\
OpenIE 4 & WIKI & 316 & 208 & 0.66 \\
OpenIE 4 & SCI & 340 & 160 & 0.47\\
\end{tabular}
\caption{Results of triples extracted from the SCI and WIKI corpora using the Open IE and MinIE tools. \label{aggresults}}
\end{center}
\label{default}
\end{table}

\begin{figure}[h]
\centering
\includegraphics[width=.9\textwidth]{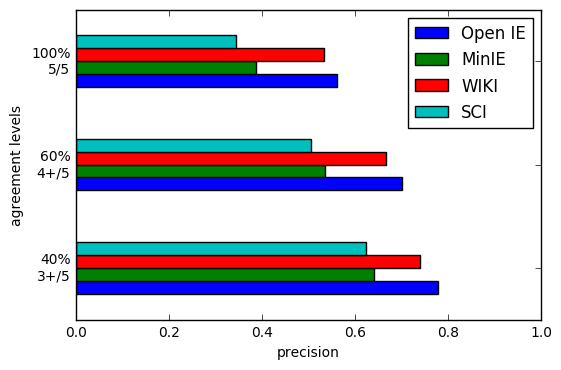}
\caption{Precision at various agreement levels. Agreement levels are shown as the proportion of overall agreement. In addition, we indicate the the minimum number of annotators who considered relations correct out of the total number of annotators.  \label{fig:agreementlevels}}
\end{figure}

Figure \ref{fig:agreementlevels} shows the performance of extractors in terms of precision as inter-annotator agreement decreases. In this figure, we look only at agreement on triples where the majority agree that the triple is correct. Furthermore, to ease comparison, we only consider triples with 5 judgements this excludes 9 triples. We indicate not only the pair-wise inter-annotator agreement but also the number of annotators who have judged a triple to be correct. For example, at the 40\% agreement level at least 3 annotators have agreed that a triple is true. The figure separates the results by extractor and by data source. 

We see that as expected the amount of triples agreed to as correct grows larger as we relax the requirement for agreement. For example, analyzing Open IE's results, at the 100\% agreement level we see a precision of 0.56 whereas at the 40\% agreement level we see a precision of 0.78. Table \ref{table:agreementnumbers} shows the total number of correct extractions at the three agreement levels. 

\begin{table}[h]
\begin{center}
\begin{tabular}{r r r r r r r}
\textsc{Agreement} & \textsc{Open IE 4} & \textsc{MinIE} & \textsc{Open IE 4} & \textsc{MinIE} & \textsc{Open IE 4} & \textsc{MinIE } \\
& \textsc{SCI} & \textsc{SCI} & \textsc{WIKI} & \textsc{WIKI} & \textsc{Total} & \textsc{Total} \\
\hline
40\%  (3+/5)  & 253  & 462 & 256 &  553 & 509 & 1015 \\
60\%	 (4+/5) & 218  & 361 & 241 & 488 & 459 & 849 \\
100\% (5/5) & 160  & 235 & 207 & 377 & 367 & 612 \\
\end{tabular}
\caption{Correct triples at different levels of agreement subsetted by system and data source \label{table:agreementnumbers}. Agreement levels follow from Figure \ref{fig:agreementlevels}}
\end{center}
\end{table}

\subsection{Testing H1: Comparing the Performance of OIE on Scientific vs. Encyclopedic Text}
From the data, we see that extractors perform better on sentences from Wikipedia (0.54 P) than scientific text (0.34 P). Additionally, we see that there is higher annotator agreement on whether triples extracted from Wikipedia and scientific text are correct or incorrect: 0.80 - SD 0.24 (WIKI)  vs. 0.72 - SD 0.25 (SCI). A similar difference in agreement is observed when only looking at triples that are considered to be correct by the majority of annotators: 0.87 - SD 0.21 (WIKI) vs. 0.78 - SD 0.25 (SCI) . In both cases, the difference is significant with p-values $<$  0.01 using Welch's t-test. The differences between data sources are also seen when looking at the individual extraction tools. For instance, for Open IE 4 the precision is 0.19 higher for wikipedia extractions over those from scientific text. {\it With this evidence, we reject our first hypothesis that the performance of these extractors are similar across data sources.}

\subsection{Testing H2: Comparing the Performance of Systems }
We also compare the output of the two extractors. In terms precision, Open IE 4  performs much better across the two datasets (0.56P vs 0.39P).  Looking at triples considered to be correct by the majority of annotators, we see that Open IE 4 has higher inter-annotator agreement 0.87 - SD 0.22 (Open IE) vs 0.81 - SD 0.24 (MinIE). Focusing on scientific and medical text (SCI), again where the triples are majority annotated as being correct, Open IE has higher inter-annotator agreement (Open IE: 0.83 - SD 0.24  vs MiniIE: 0.76 - SD 0.25). In both cases, the difference is significant with p-values $<$  0.01 using Welch's t-test. This leads us to conclude that Open IE produces triples that annotators are more likely to agree as being correct. 

MinIE provides many more correct extractions than OpenIE 4 (935 more across both datasets). The true recall numbers of the two systems can not be calculated with the data available, but the 40\% difference in the numbers of correct extractions is strong evidence that the two systems do not have equivalent behavior. 

A third indication of differences in their outputs comes from examining the complexity of the extracted relations. Open IE 4 generates longer triples on average (11.5 words) vs. 8.5 words for MinIE across all argument positions. However, Open IE 4 generates shorter relation types than MinIE (Open IE - 3.7 words; MiniIE 6.27 words) and the standard deviation in terms of word length is much more compact for Open IE 4 - ~1 word vs ~3 words for MinIE. {\it Overall, our conclusion is that Open IE 4 performs better than MinIE both in terms of precision and compactness of relation types, while not matching MinIE's recall, and thus we reject our second hypothesis. }

\subsection{Other Observations}
The amount of triples extracted from the scientific text is slightly larger than that extracted from the Wikipedia text. This follows from the fact that the scientific sentences are on average roughly 7 words longer than encyclopedic text. 

The results of our experiment also confirm the notion that an unsupervised approach to extracting relations is important. We have identified 698 unique relation types that are part of triples agreed to be correct by all annotators. This number of relation types is derived from only ~400 sentences. While not every relation type is essential for downstream tasks, it is clear that building specific extractors for each relation type in a supervised setting would be difficult. 

\section{Error Analysis}
We now look more closely at the various errors that were generated by the two extractors. 

Table \ref{notriples} shows the sentences in which neither extractor produced triples. We see 3 distinct groups. The first are phrases that are incomplete sentences usually originating from headings (e.g. Materials and methods). The next group are descriptive headings potentially coming from paper titles or figure captions. We also see a group with more complex prepositional phrases. In general, these errors could be avoided by being more selective of the sentences used for random selection. Additionally, these systems could look at potentially just extracting noun phrases with variable relation types, hence, expressing a cooccurrence relation. 

\begin{table}
\begin{tabular}{l p{13.5cm}}
\hline
\textsc{Source} & \textsc{Sentence} \\
\hline
\\
\cline{2-2} SCI & Note that Eq. \\
& Materials and methods \\
& Site and experimental carbonate chemistry \\
& Soil aggregate characteristics \\
& An equation analogous to Eq. \\
\cline{2-2}  & An Experimental Platform for the Efficient Generation of Human Cranial Placodes In Vitro \\
& Analysis 1: Manifest Vascular and Nonvascular Disease as a Predictor of Depressive Symptoms \\
& Autografts Elicit Only a Minimal Immune Response in the Primate Brain \\
& Production of wild-type TTR and ATTR Val30Met by differentiated hepatocyte-like cells \\
& Cryopreservation and recovery of hiPSCs in suspension culture \\
\cline{2-2} & For a detailed description of analysis methods and precision see Lee et al. (1997). \\
& Thus there are, up to associativity, only finitely many such q. \\
& In the absence of heavy elements, H3+ forms near the base of the model and subsequent infrared cooling balances the EUV heating rates. \\
\cline{2-2} & Assume that L is a line bundle. \\
& The set of the representative points is called the non-dominated front (or Pareto front). \\
\\
WIKI & Simultaneously won the `` Hope of the World Ballet '' Prize . \\
\\
\hline
\end{tabular}
\caption{Sentences in which no triples were extracted \label{notriples}}
\end{table}
% grep for finiding unique articles
%grep -h -E 'S[0-9A-Z]+:' *.txt | sort | uniq | wc -l
%grep -h -E 'wiki' *.txt | sort | uniq | wc -l
	
We also looked at where there was complete agreement by all annotators that a triple extraction was incorrect. In total there were 138 of these triples originating from 76 unique sentences. There were several patterns that appeared in these sentences. 
\begin{itemize}
\item Long complex sentences led to incorrect extractions. For example, {\tt \small ``Most strikingly, there was a mutant gene-dose-dependent increase in caspase-cleaved TAU fragments, as determined by the caspase-cleaved TAU-specific antibody C3 (Gamblin et al., 2003; Guillozet-Bongaarts et al., 2005), in neurons derived from the isogenic TAU-A152T-iPSCs (Figures 4J and 4K).''}) led to the following triples that were deemed incorrect: {\tt isogenic TAU-A152T-iPSCs :: is :: Figures}  and {\tt Gamblin et :: is :: caspase-cleaved TAU-specific antibody C3}
\item The use of pronouns as subjects led to triples that were deemed to be incorrect. (e.g. {\tt \small ``So he was forced to requisition not only the public treasury of Gades but also the wealth from its temples.} led to the following incorrect triples:  {\tt he :: was forced :: to requisition not only the public treasury of Gades but also the wealth from its temples} and  {\tt he :: to requisition :: not only the public treasury of Gades but also the wealth from its temples}.
\item Complex mathematical formula often generated incorrect triples
\item MinIE as part of its extraction process substitutes quantities with variables (e.g. QUANT\_1). We included this potential in our labelling instructions but this often led to triples that were labelled as incorrect by the annotators. We believe this source of error could stem from the given instructions. 
\item Reuse of abbreviations as adjectives also led to incorrect triples. For example, ``BMP antagonists'' and ``BMP pathway'' in the following sentence: {\tt We also observed significant transcriptional changes in WNT and BMP pathway components such as an increase in the WNT pathway inhibitor DKK-1 and BMP antagonists, such as GREMLIN-1 and BAMBI (Figures 2B-2D), which are known transcriptional targets of BMP signaling (Grotewold et al., 2001).} led to the following triples that were labelled incorrect:  {\tt BMP antagonists :: are known ::}
and {\tt increase :: is :: bmp pathway component}.
\end{itemize}

We also see similar errors to those pointed out by \cite{Schneider2017}, namely, uninformative extractions, the difficulty in handling n-ary relations that are latent in the text, difficulties handling negations, and very large argument lengths. In general, these errors together point to several areas for further improvement including:
\begin{itemize}
\item deeper co-reference resolution either for variables in mathematical formula or for pronouns;
\item improved handling of prepositional phrases;
\item relaxing requirements for correct grammar within sentences;
\item better handling of abbreviations. 
\end{itemize}

\section{Conclusion}
The pace of change in the scientific literature means that interconnections and facts in the form of relations between entities are constantly being created. Open information extraction provides an important tool to keep up with that pace of change. We have provided evidence that unsupervised techniques are needed to be able to deal with the variety of relations present in text. The work presented here provides an independent evaluation of these tools in their use on scientific text. Past evaluations have focused on encyclopedic or news corpora which often have simpler structures.  We have shown that existing OIE systems perform worse on scientific and medical content than on general audience content.  

There are a range of avenues for future work. First, the application of Crowd Truth framework~\cite{aroyo2013crowd} in the analysis of these results might prove to be useful as we believe that the use of unanimous agreement tends to negatively impact the perceived performance of the OIE tools. Second, we think the application to n-ary relations and a deeper analysis of negative relations would be of interest. To do this kind of evaluation, an important area of future work is the development of guidelines and tasks for more complex analysis of sentences in a crowd sourcing environment. The ability, for example, to indicate argument boundaries or correct sentences can be expected of expert annotators but needs to implemented in a manner that is efficient and easy for the general crowd worker. Third, we would like to expand the evaluation dataset to an even larger numbers of sentences.  Lastly, there are a number of core natural language processing components that might be useful for OIE in this setting, for example, the use of syntactic features as suggested by~\cite{Christensen2011}. Furthermore, we think that coreference is a crucial missing component and we are actively investigating improved coreference resolution for scientific texts. 

To conclude, we hope that this evaluation provides further insights for implementors of these extraction tools to deal with the complexity of scientific and medical text.

\bibliographystyle{acl}
\bibliography{references}

\begin{thebibliography}{}

\bibitem[\protect\citename{Aroyo and Welty}2013]{aroyo2013crowd}
Lora Aroyo and Chris Welty.
\newblock 2013.
\newblock Crowd truth: Harnessing disagreement in crowdsourcing a relation
  extraction gold standard.
\newblock {\em WebSci2013. ACM}, 2013.

\bibitem[\protect\citename{Artstein and Poesio}2008]{artstein2008inter}
Ron Artstein and Massimo Poesio.
\newblock 2008.
\newblock Inter-coder agreement for computational linguistics.
\newblock {\em Computational Linguistics}, 34(4):555--596.

\bibitem[\protect\citename{Banko \bgroup et al.\egroup }2007]{Banko2007}
Michele Banko, Michael~J. Cafarella, Stephen Soderland, Matt Broadhead, and
  Oren Etzioni.
\newblock 2007.
\newblock Open information extraction from the web.
\newblock In {\em Proceedings of the 20th International Joint Conference on
  Artifical Intelligence}, IJCAI'07, pages 2670--2676, San Francisco, CA, USA.
  Morgan Kaufmann Publishers Inc.

\bibitem[\protect\citename{Bornmann and Mutz}2015]{bornmann2015growth}
Lutz Bornmann and R{\"u}diger Mutz.
\newblock 2015.
\newblock Growth rates of modern science: A bibliometric analysis based on the
  number of publications and cited references.
\newblock {\em Journal of the Association for Information Science and
  Technology}, 66(11):2215--2222.

\bibitem[\protect\citename{Christensen \bgroup et al.\egroup
  }2011]{Christensen2011}
Janara Christensen, Mausam, Stephen Soderland, and Oren Etzioni.
\newblock 2011.
\newblock An analysis of open information extraction based on semantic role
  labeling.
\newblock In {\em Proceedings of the Sixth International Conference on
  Knowledge Capture}, K-CAP '11, pages 113--120, New York, NY, USA. ACM.

\bibitem[\protect\citename{Crump \bgroup et al.\egroup
  }2013]{10.1371/journal.pone.0057410}
Matthew J.~C. Crump, John~V. McDonnell, and Todd~M. Gureckis.
\newblock 2013.
\newblock Evaluating amazon's mechanical turk as a tool for experimental
  behavioral research.
\newblock {\em PLOS ONE}, 8(3):1--18, 03.

\bibitem[\protect\citename{Del~Corro and Gemulla}2013]{Corro2013}
Luciano Del~Corro and Rainer Gemulla.
\newblock 2013.
\newblock Clausie: Clause-based open information extraction.
\newblock In {\em Proceedings of the 22Nd International Conference on World
  Wide Web}, WWW '13, pages 355--366, New York, NY, USA. ACM.

\bibitem[\protect\citename{Erlewine and Kotek}2016]{Erlewine2016}
Michael~Yoshitaka Erlewine and Hadas Kotek.
\newblock 2016.
\newblock {A streamlined approach to online linguistic surveys}.
\newblock {\em Natural Language {\&} Linguistic Theory}, 34(2):481--495, may.

\bibitem[\protect\citename{Fader \bgroup et al.\egroup }2011]{Fader2011}
Anthony Fader, Stephen Soderland, and Oren Etzioni.
\newblock 2011.
\newblock Identifying relations for open information extraction.
\newblock In {\em Proceedings of the Conference on Empirical Methods in Natural
  Language Processing}, EMNLP '11, pages 1535--1545, Stroudsburg, PA, USA.
  Association for Computational Linguistics.

\bibitem[\protect\citename{Fort \bgroup et al.\egroup }2011]{Fort2011}
Kar{\"{e}}n Fort, Gilles Adda, and K.~Bretonnel Cohen.
\newblock 2011.
\newblock {Amazon Mechanical Turk: Gold Mine or Coal Mine?}
\newblock {\em Computational Linguistics}, 37(2):413--420, jun.

\bibitem[\protect\citename{Gashteovski \bgroup et al.\egroup
  }2017]{Gashteovski2017}
Kiril Gashteovski, Rainer Gemulla, and Luciano {Del Corro}.
\newblock 2017.
\newblock {MinIE: Minimizing Facts in Open Information Extraction}.
\newblock {\em Proceedings of the 2017 Conference on Empirical Methods in
  Natural Language Processing}, pages 2620--2630.

\bibitem[\protect\citename{Groth \bgroup et al.\egroup
  }2016]{DBLP:conf/akbc/GrothPMAD16}
Paul~T. Groth, Sujit Pal, Darin McBeath, Brad Allen, and Ron Daniel.
\newblock 2016.
\newblock Applying universal schemas for domain specific ontology expansion.
\newblock In Jay Pujara, Tim Rockt{\"{a}}schel, Danqi Chen, and Sameer Singh,
  editors, {\em Proceedings of the 5th Workshop on Automated Knowledge Base
  Construction, AKBC@NAACL-HLT 2016, San Diego, CA, USA, June 17, 2016}, pages
  81--85. The Association for Computer Linguistics.

\bibitem[\protect\citename{Leveling}2015]{leveling2015tagging}
Johannes Leveling.
\newblock 2015.
\newblock Tagging of temporal expressions and geological features in scientific
  articles.
\newblock In {\em Proceedings of the 9th Workshop on Geographic Information
  Retrieval}, GIR '15, pages 6:1--6:10, New York, NY, USA. ACM.

\bibitem[\protect\citename{Mausam}2016]{Mausam2016}
Mausam Mausam.
\newblock 2016.
\newblock Open information extraction systems and downstream applications.
\newblock In {\em Proceedings of the Twenty-Fifth International Joint
  Conference on Artificial Intelligence}, IJCAI'16, pages 4074--4077. AAAI
  Press.

\bibitem[\protect\citename{Poon \bgroup et al.\egroup
  }2014]{distant-supervision-cancer-pathway-extraction-text}
Hoifung Poon, Kristina Toutanova, and Chris Quirk.
\newblock 2014.
\newblock Distant supervision for cancer pathway extraction from text.
\newblock In {\em Pacific Symposium on Biocomputing Co-Chairs}, pages 120--131.
  World Scientific.

\bibitem[\protect\citename{Quirk and
  Poon}2017]{distant-supervision-relation-extraction-beyond-sentence-boundary}
Chris Quirk and Hoifung Poon.
\newblock 2017.
\newblock Distant supervision for relation extraction beyond the sentence
  boundary.
\newblock In Mirella Lapata, Phil Blunsom, and Alexander Koller, editors, {\em
  Proceedings of the 15th Conference of the European Chapter of the Association
  for Computational Linguistics, {EACL} 2017, Valencia, Spain, April 3-7, 2017,
  Volume 1: Long Papers}, pages 1171--1182. Association for Computational
  Linguistics.

\bibitem[\protect\citename{Schneider \bgroup et al.\egroup
  }2017]{Schneider2017}
Rudolf Schneider, Tom Oberhauser, Tobias Klatt, Felix~A. Gers, and Alexander
  L{\"o}ser.
\newblock 2017.
\newblock Analysing errors of open information extraction systems.
\newblock In {\em Proceedings of the First Workshop on Building Linguistically
  Generalizable NLP Systems}, pages 11--18. Association for Computational
  Linguistics.

\end{thebibliography}

\end{document}